\documentclass[11pt]{article}

\usepackage{latexsym}
\usepackage{amsmath}
\usepackage{bm}
\usepackage{amsthm}
\usepackage{color}
\usepackage{amssymb}
\usepackage{url}
\usepackage[latin1]{inputenc}
\usepackage{alltt}
\usepackage{verbatim}
\usepackage{graphicx}
\usepackage{xspace}
\usepackage{subfig}

\usepackage{natbib}
\usepackage{algorithm}
\usepackage[noend]{algorithmic}

\usepackage{datetime}

\newcommand{\ra}{\ensuremath{\rightarrow}\xspace}
\newcommand{\la}{\ensuremath{\leftarrow}\xspace}

\newcommand{\mdc}[1]{}

\newcommand{\mdcomment}[1]{}

\newcommand{\mymarginpar}[1]{}

\newcommand{\Astar}{A\ensuremath{^{*}}\xspace}

\newcommand{\p}{\ensuremath{p}\xspace}
\newcommand{\q}{\ensuremath{q}\xspace}
\newcommand{\pbar}{\bar{\p}\xspace}
\newcommand{\qbar}{\bar{\q}\xspace}

\floatstyle{ruled}
\newfloat{algo}{thp}{lop}
\floatname{algo}{Algorithm}

\newcommand{\OSstar}{OS\ensuremath{^{*}}\xspace}

\newcommand{\norm}[1]{\|#1\|}

\newenvironment{changedEnv}{}{}

\newcommand{\changed}[1]{#1}

\usepackage{authblk}

\title{The \OSstar algorithm: a Joint approach to Exact Optimization and Sampling\\ {\normalsize{**** Version 0.9 ****}}\\ } 

\author[1]{Marc Dymetman}
\author[1]{Guillaume Bouchard}
\author[2]{Simon Carter\footnote{Work conducted during an internsphip at XRCE.}\,}
\affil[1]{ Xerox Research Centre Europe\authorcr 6, chemin de Maupertuis\authorcr 38240 Meylan, France\authorcr \tt{first.last@xrce.xerox.com}\authorcr\ }
\affil[2]{ ISLA, University of Amsterdam, Science Park 904\authorcr 1098 XH Amsterdam, The Netherlands\authorcr \tt{s.c.carter@uva.nl}}

\begin{document} 
 \maketitle 




\begin{abstract} 
Most current sampling algorithms for high-dimensional distributions are based on MCMC techniques and are approximate in the sense that they are valid only asymptotically. Rejection sampling, on the other hand, produces valid samples, but is unrealistically slow in high-dimension spaces. The OS* algorithm that we propose is a unified approach to exact optimization and sampling, based on incremental refinements of a functional upper bound, which combines ideas of adaptive rejection sampling and of A* optimization search. We show that the choice of the refinement can be done in a way that ensures tractability in high-dimension spaces, and we present first experiments in two different settings: inference in high-order HMMs and in large discrete graphical models.\\[1cm]
\end{abstract}

\section{Introduction}

Common algorithms for sampling high-dimensional distributions are based on MCMC techniques \citep{Andrieu2003,Robert2004}, which are approximate in the sense that they produce valid samples only asymptotically. By contrast, the elementary technique of Rejection Sampling \citep{Robert2004} directly produces exact samples, but, if applied naively to high-dimensional spaces, typically requires unacceptable time before producing a first sample.

The algorithm that we propose, \OSstar, is a joint exact \underline{O}ptimization and \underline{S}ampling algorithm that is inspired both by rejection sampling and by classical A\underline{\mbox{\(^{*}\)}} optimization, and which can be applied to high-dimensional spaces. The main idea is to upper-bound the complex target distribution $p$ by a simpler proposal distribution $q$, such that a dynamic programming (or another low-complexity) method can be applied to $q$ in order to efficiently sample or maximize from it. In the case of sampling, rejection sampling is then applied to $q$, and on a reject at point $x$, $q$ is refined into a slightly more complex $q'$ in an adaptive way. This is done by using the evidence of the reject at $x$, implying a gap between $q(x)$ and $p(x)$, to identify a (soft) constraint implicit in $p$ which is not accounted for by $q$, and by integrating this constraint in $q$ to obtain $q'$.

The constraint which is integrated tends to be highly relevant and to increase the acceptance rate of the algorithm. By contrast, many constraints that are constitutive of $p$ are never ``activated'' by sampling from $q$, because $q$ never explores regions where they would become visible. For example, anticipating on our HMM experiments in section \ref{sec:hmms}, there is little point in explicitly including in $q$ a 5-gram constraint on a certain latent sequence in the HMM if this sequence is already unlikely at the bigram level: the bigram constraints present in the proposal $q$ will ensure that this sequence will never (or very rarely) be explored by $q$.

The case of optimization is treated in exactly the same way as sampling. Formally, this consists in moving from assessing proposals in terms of the $L_1$ norm to assessing them in terms of the $L_\infty$ norm. Typically, when a dynamic programming procedure is available for sampling ($L_1$ norm) with $q$, it is also available for maximizing from $q$ ($L_\infty$ norm), and the main difference between the two cases is then in the criteria for selecting among possible refinements.

\paragraph{Related work}
In an heuristic optimization context the two interesting, but apparently little known, papers \citep{Kam1996,did-lin} , discuss a technique for decoding images based on high-order language models for which upper-bounds are constructed in terms of simpler variable-order models. Our  application of \OSstar in section \ref{sec:hmms} to the problem of maximizing a high-order HMM is similar to their technique; however \citep{Kam1996,did-lin}  do not attempt to generalize their approach to other optimization problems amenable to dynamic programming or discuss any connection to sampling.

In order to improve the acceptance rate of rejection sampling, one has to lower the proposal $q$ curve as much as possible while keeping it above the $p$ curve. In order to do that, some authors \citep{Gilks1992,Gorur2008}, have proposed \emph{Adaptive Rejection Sampling  (ARS)} where, based on rejections, the $q$ curve is updated to a lower curve $q'$ with a better acceptance rate. These techniques have predominantly been applied to continuous distributions on the one-dimensional real line where convexity assumptions on the target distribution can be exploited to progressively approximate it tighter and tighter through upper bounds consisting of piecewise linear envelopes.

Also in the context of rejection sampling, \citep{Mansinghka2009}  considers the case of a probabilistic graphical model; it introduces an heuristically determined order of the variables in this model and uses this (fixed) order to define a sequence of exact samplers over an increasing set of variables, where the exact sampler over the first $k+1$ variables is recursively obtained by using the preceding exact sampler over the first $k$ variables and accepting or rejecting its samples based on the $(k+1)^{\text{th}}$ variable. While our experiments on graphical models in section \ref{sec:ExperimentsGraphical} have some similarities to this approach, they do not use a cascade of exact samplers, but rather they partition the space of configurations dynamically based on rejects experienced by the current proposal.

\section{The OS* algorithm}
\label{sec:algorithm}

\OSstar is a unified algorithm for optimization and sampling. For simplicity, we first present its sampling version, then move to its optimization version, and finally get to the unified view. We start with some background and notations about rejection sampling.

\subsection{Background}
\label{sec:background}


Let $\p:X\rightarrow \mathbb{R}_{+}$ be a measurable $L_1$ function with respect to a base measure $\mu$ on a space $X$, i.e. 
$\int_X p(x) d\mu(x) < \infty$.  We define $\pbar(x) \equiv \frac{\p(x)}{\int_X p(x) d\mu(x)}$. The function \p can be seen as an unnormalized density over $X$, and $\pbar$ as a normalized density which defines a probability distribution over $X$, called the \emph{target distribution}, from which we 
want to sample from\footnote{By abuse of language, we will also say that a sample from $\pbar$ is a sample from $\p$.}. While we may not be able to sample directly from the target distribution $\pbar$, let us assume that we can easily compute $p(x)$ for any given $x$. 
\emph{Rejection Sampling} (RS) \citep{Robert2004} then works as follows. We define a certain unnormalized \emph{proposal density} $\q$ over $X$, which is such that (i) we know how to directly sample from it (more precisely, from $\qbar$), and (ii) $q$ dominates $p$, i.e. for all $x\in X, p(x) \leq q(x)$. We then repeat the following process: (1) we draw a sample $x$ from $q$, (2) we compute the ratio $r(x)\equiv p(x)/q(x) \leq 1$, (3) with probability $r(x)$ we accept $x$, otherwise we reject it, (4) we repeat the process with a new $x$. It can then be shown that this procedure produces an exact sample from $\p$. Furthermore, the average rate at which it produces these samples, the \emph{acceptance rate}, is equal to $P(X)/Q(X)$ \citep{Robert2004}, where for a (measurable) subset $A$ of $X$, we define $P(A) \equiv \int_A p(x) d\mu(x)$ and similarly with $Q$. In Fig.~\ref{fig:panel1}, panel (S1), the acceptance rate is equal to the ratio of the area below the $p$ curve with that below the $q$ curve.



\subsection{Sampling with \OSstar}
\begin{figure*}
\includegraphics[clip=false, draft=false, trim=0cm 3cm 4cm 8cm, scale=.5]{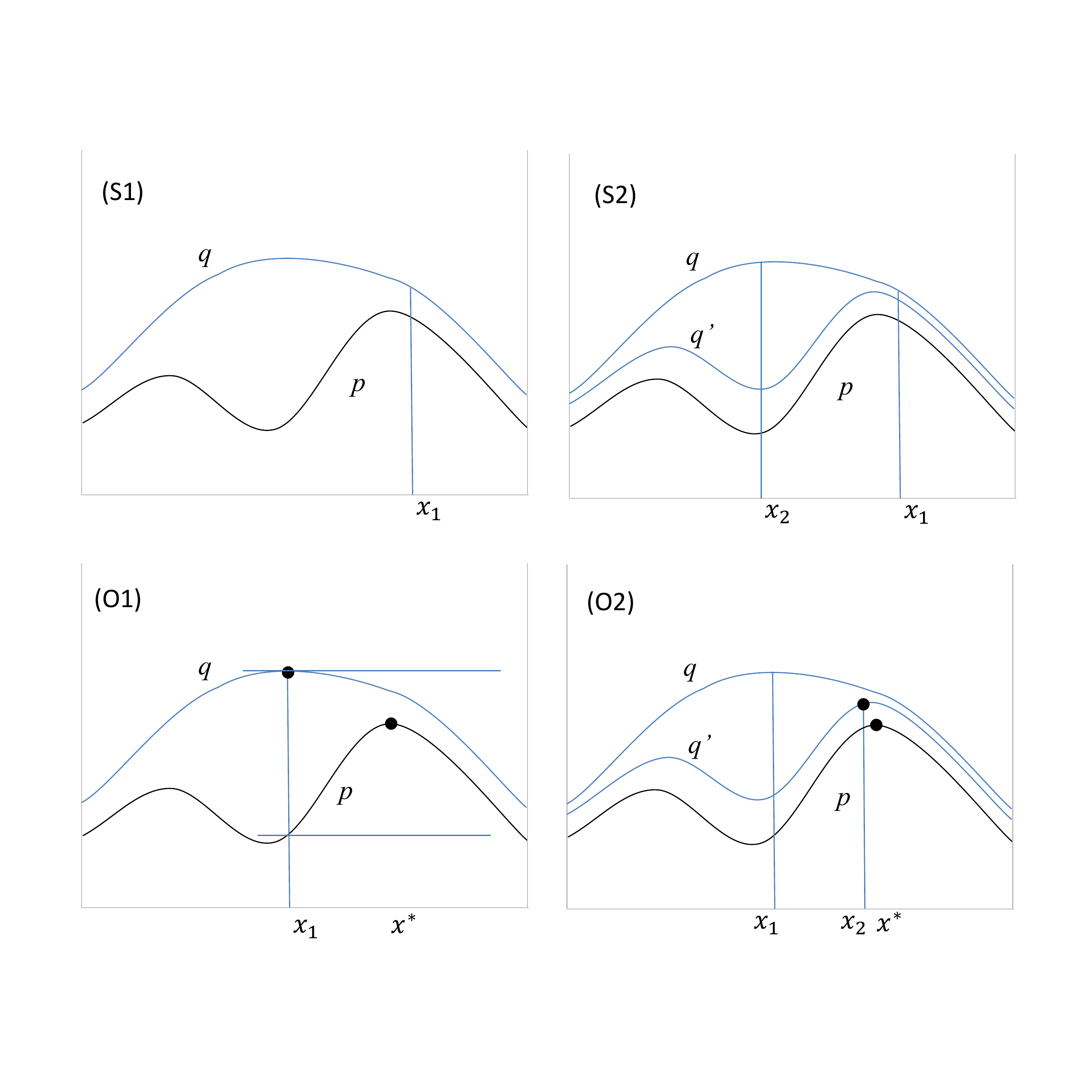}
\caption{Sampling with \OSstar (S1, S2), and optimization with \OSstar (O1, O2).}
	\label{fig:panel1}
\end{figure*}

The way \OSstar does sampling is illustrated on the top of Fig.~\ref{fig:panel1}. In this illustration, we start sampling with an initial proposal density $q$ (see (S1)). Our first attempt produces $x_1$, for which the ratio $r_{q}(x_1) = p(x_1)/q(x_1)$ is close to $1$; this leads, say, to the acceptance of $x_1$. Our second attempt produces $x_2$, for which the ratio $r_{q}(x_2) = p(x_2)/q(x_2)$ is much lower than $1$, leading, say, to a rejection. Although we have a rejection, we have gained some useful information, namely that $p(x_2)$ is much lower than $q(x_2)$, and we are going to use that ``evidence'' to define a new proposal $q'$ (see (S2)), which has the following properties:
\begin{itemize}
\item One has $p(x) \leq q'(x) \leq q(x)$ everywhere on $X$.
\item One has $q'(x_2) < q(x_2)$.
\end{itemize}
One extreme way of obtaining such a $q'$ is to take:
$$
q'(x) \equiv
\begin{cases} 
p(x) & \text{if } x=x_{2}\\
q(x) & \text{if } x\neq x_{2}
\end{cases}
$$ 
which, when the space $X$ is discrete,
has the effect of improving the acceptance rate, but only slightly so, by insuring that any time $q'$ happens to select $x_2$, it will accept it.

A better generic way to find a $q'$ is the following. Suppose that we are provided with a small finite set of ``one-step refinement actions" $a_j$, depending on $q$ and $x_2$, which are able to move from $q$ to a new $q_j' = a_j(q,x_2)$ such that for any such $a_j$ one has  $p(x) \leq q_j'(x) \leq q(x)$ everywhere on $X$ and also $q_j'(x_2) < q(x_2)$. Then we will select among these $a_j$ moves the one that is such that the $L_1$ norm of $q_j'$ is minimal among the possible $j$'s, or in other words, such that $\int_X q_j'(x) d\mu(x)$ is minimal in $j$. The idea there is that, by doing so, we will improve the acceptance rate of $q'_j$  (which depends directly on $\|q_j'\|_1$) as much as possible, while (i) not having to explore a too large space of possible refinements, and (ii) moving from a representation for $q$ to an only slightly more complex representation for $q'_j$, rather than to a much more complex representation for a $q'$ that could result from exploring a larger space of possible refinements for $q$.\footnote{In particular, even if we could find a refinement $q'$ that would exactly coincide with $p$, and therefore would have the smallest possible $L_1$ norm, we might not want to use such a refinement if this involved an overly complex representation for $q'$.}
The intuition behind such one-step refinement actions $a_j$ will become clearer when we consider concrete examples later in this paper.\mdcomment{The intuition could be made more formal by explaining the stuff about exponential-family models with some active features and a lot of inactive features. However this would take some care and also some space.}

\subsection{Optimization with \OSstar}
The optimization version of \OSstar is illustrated on the bottom of Fig.~\ref{fig:panel1}, where (O1) shows on the one hand the function $p$ that we are trying to maximize from, along with its (unknown) maximum $p^{*}$, indicated by a black circle on the $p$ curve, and corresponding to $x^{*}$ in $X$. It also shows a ``proposal'' function $q$ which is such --- analogously to the sampling case --- that (1) the function $q$ is above $p$ on the whole of the space $X$ and (2) it is easy to directly find the point $x_1$ in $X$ at which it reaches its maximum $q^{*}$, shown as a black circle on the $q$ curve.


A simple, but central, observation is the following one. Suppose that the distance between $q(x_1)$ and $p(x_1)$ is smaller than $\epsilon$, then the distance between $q(x_1)$ and $p^{*}$ is also smaller than $\epsilon$. This can be checked immediately on the figure, and is due to the fact that on the one hand $p^{*}$ is higher than $p(x_1)$, and that on the other hand it is below $q(x^{*})$, and \emph{a fortiori} below $q(x_1)$. In other words, if the maximum that we have found for $q$ is at a coordinate $x_1$ and we observe that $q(x_1)-p(x_1) < \epsilon$, then we can conclude that we have found the maximum of $p$ up to $\epsilon$.

In the case of $x_1$ in the figure, we are still far from the maximum, and so we ``reject'' $x_1$, and refine $q$ into $q'$ (see (O2)),  using exactly the same approach as in the sampling case, but for one difference: the one-step refinement option $a_j$ that is selected is now chosen on the basis of how much it decreases, not the $L_1$ norm of $q$, but the max of $q$ --- where, as a reminder, this max can also be notated $\|q\|_\infty$, using the $L_\infty$ norm notation.\footnote{A formal definition of that norm is that $\|q\|_\infty$ is equal to the ``essential supremum'' of $q$ over $(X,\mu)$ (see below), 
but for all practical purposes here, it is enough to think of this essential supremum as being the max, when it exists.}

Once this $q'$ has been selected, one can then find its maximum at $x_2$ and then the process can be repeated with $q_{1}=q,  q_{2}=q', ...$ until the difference between $q_{k}(x_k)$ and $p(x_k)$ is smaller than a certain predefined threshold.\mdc{Show the equivalence between multiplicative and subtractive thresholds.}

\subsection{Sampling $L_1$ vs. Optimization $L_\infty$}
While sampling and optimization are usually seen as two completely distinct tasks, they can actually be viewed as two extremities of a continuous range, when considered in the context of $L_p$ spaces \citep{Ash1999}. 

If $(X, \mu)$ is a measure space, and if $f$ is a real-valued function on this space, one defines the $L_p$ norm $\norm{f}_p$, for $1 \leq p < \infty$ as:
$$
\norm{f}_p \equiv \left(\int_X |f|^p(x) d\mu(x)\right)^{1/p}.
$$
One also defines the the $L_\infty$ norm $\norm{f}_\infty$ as:
$$
\norm{f}_\infty \equiv \inf \{ C\ge 0 : |f(x)| \le C \mbox{ for almost every } x\},
$$
where the right term is called the \emph{essential supremum} of $|f|$, and can be thought of roughly as the ``max'' of the function. So, with some abuse of language, we can simply write:
$
\norm{f}_\infty \equiv \max_{x\in X}|f| .
$
The space $L_p$, for $1 \leq p \leq \infty$, is then defined as being the space of all functions $f$ for which $\norm{f}_p  < \infty$.

Under the simple condition that $\norm{f}_p < \infty$ for some $p < \infty$, we have:
$
\lim_{p \ra \infty}  \norm{f}_p = \norm{f}_\infty.
$

The standard notion of sampling is relative to $L_1$. However we can introduce the following generalization --- where we use the notation $L_\alpha$ instead of $L_p$ in order to avoid confusion with our use of $p$ for denoting the target distribution. We will say that we are performing \emph{sampling of a non-negative function $f$ relative to $L_\alpha(X,\mu)$}, for $1\leq \alpha <\infty$,  if $f \in L_\alpha(X,\mu)$ and if we sample --- in the standard sense --- according to the normalized density distribution $\bar{f}(x) \equiv \frac{f(x)^\alpha}{\int_X f(x)^\alpha d\mu(x)}$. In the case $\alpha = \infty$, we will say that we are sampling relative to $L_\infty(X,\mu)$, if $f \in L_\infty(X,\mu)$ and if we are performing optimization relative to $f$, more precisely, if for any $\epsilon > 0$, we are able to find an $x$ such that $|\norm{f}_\infty-f(x)| < \epsilon$.\mdc{Check this definition.}

Informally, sampling relative to $L_\alpha$ ``tends'' to sampling with $L_\infty$ (i.e. optimization), for $\alpha$ tending to $\infty$, in the sense that for a large $\alpha$, an $x$ sampled relative to $L_\alpha$ ``tends'' to be close to a maximum for $f$. We will not attempt to give a precise formal meaning to that observation here, but just note the connection with the idea of \emph{simulated annealing} \citep{Kirkpatrick1983}, which we can view as a mix between the MCMC Metropolis-Hastings sampling technique \citep{Robert2004} and the idea of sampling in $L_\alpha$ spaces with larger and larger $\alpha$'s.

In summary, we thus can view optimization as an extreme form of sampling. In the sequel we will often use this generalized sense of sampling in our algorithms.\footnote{Note: While our two experiments in section \ref{sec:experiments} are based on discrete spaces, the \OSstar algorithm is more general, and \emph{can be applied to any measurable space (in particular continuous spaces)}; in such cases, $p$ and $q$ have to be measurable functions, and the relation $p \leq q$ should be read as $p(x) \leq q(x)$ a.e. (almost everywhere) relative to the base measure $\mu$.}

\subsection{\OSstar as a unified algorithm}
The general design of \OSstar can be described as follows:
\begin{itemize}
\item Our goal is to OS-sample from $p$, where we take the expression ``OS-sample'' to refer to a generalized sense that covers both sampling (in the standard sense) and optimization.
\item We have at our disposal a family $\mathcal{Q}$ of proposal densities over the space $(X,\mu)$, such that, for every $q\in \mathcal{Q}$, we are able to OS-sample efficiently from $q$.
\item Given a reject $x_1$ relative to a proposal $q$, with $p(x_1) < q(x_1)$, we have at our disposal a (limited) number of possible ``one-step'' refinement options $q'$, with $p \leq q' \leq q$, and such that $q'(x_1) < q(x_1)$.
\item We then select one such $q'$. One possible selection criterion is to prefer the $q'$ which has the smallest $L_1$ norm (sampling case) or $L_\infty$ norm (optimization). In one sense, this is the most natural criterion, as it means we are directly lowering the norm that controls the efficiency of the OS-sampling; for instance, for sampling, if $q_1'$ and $q_2'$ are two candidates refinements with $\norm{q_1'}_1 < \norm{q_2'}_1$, then the acceptance rate of $q_1'$ is larger than that of $q_2'$, simply because then $P(X)/Q_1'(X) > P(X)/Q_2'(X)$ , and similarly, in optimization, if $\norm{q_1'}_\infty < \norm{q_2'}_\infty$, then the gap between $\max_x(q_1'(x))$ and $p^*$ is smaller than that between $\max_x(q_2'(x))$ and $p^*$, simply because then $\max_x(q_1'(x)) < \max_x(q_2'(x))$. 
However, using this criterion may require the computation of the norm of each of the possible one-step refinements, which can be costly, and one can prefer simpler criteria, for instance simply selecting the $q'$ that minimizes $q'(x_1)$.\mdcomment{Maybe this should be expanded and linked to examples in the experiments ? (not sure)}

\item We iterate until we settle on a ``good'' $q$: either (in sampling) one which is such that the cumulative acceptance rate until this point is above a certain threshold; or (in optimization) one for which the ratio $p(x_1)/q(x_1)$ is closer to $1$ than a certain threshold, with $x_1$ being the maximum for $q$.
\end{itemize}

The following algorithm gives a unified view of \OSstar, valid for both sampling and optimization. This is a high-level view, with some of the content delegated to the subroutines \texttt{OS-Sample, Accept-or-Reject, Update, Refine, Stop}, which are described in the text.
\begin{algorithm}[H]
\caption{The \OSstar algorithm}
\label{algo:OSstar}
\begin{algorithmic}[1] 
\WHILE { \NOT Stop($h$) }
\STATE OS-Sample $x \sim q$ 
\STATE $r \la p(x)/q(x)$
\STATE  Accept-or-Reject$(x,r)$
\STATE  Update($h,x$)
\IF {Rejected($x$)}
\STATE $q$ \la Refine$(q,x)$
\ENDIF
\ENDWHILE
\RETURN \!\!$q$ along with accepted $x$'s in $h$
\end{algorithmic}
\end{algorithm}

On entry into the algorithm, we assume that we are either in sample mode or in optimization mode, and also that we are starting from a proposal $q$ which (1) dominates $p$ and (2) from which we can sample or optimize directly. We use the terminology \emph{OS-Sample} to represent either of these cases, where \texttt{OS-Sample $x \sim q$} refers to sampling an $x$ according to the proposal $q$ or optimizing $x$ on $q$ (namely finding an $x$ which is an argmax of $q$), depending on the case. On line (1), $h$ refers to the history of the sampling so far, namely to the set of trials $x_1, x_2, ...$ that have been done so far, each being marked for acceptance or rejection (in the case of sampling, this is the usual notion, in the case of optimization, all but the last proposal will be marked as rejects). The stopping criterion \texttt{Stop($h$)} will be to stop: 
(i) \emph{in sampling mode}, if the number of acceptances so far relative to the number of trials is larger than a certain predefined threshold, and in this case will return on line (8), first, the list of accepted $x$'s so far, which is already a valid sample from $p$, and second, the last refinement $q$, which can then be used to produce any number of future samples as desired with an acceptance ratio similar to the one observed so far;
(ii) \emph{in optimization mode}, if the last element $x$ of the history is an accept, and in this case will return on line (8), first the value $x$, which in this mode is the only accepted trial in the history, and second, the last refinement $q$ (which can be used for such purposes as providing a ``certificate of optimality of $x$ relative to $p$'', but we do not detail this here).

On line (3), we compute the ratio $r$, and then on line (4) we decide to accept $x$ or not based on this ratio; in optimization mode, we accept $x$ if the ratio is close enough to $1$, as determined by a threshold\footnote{When $X$ is a finite domain, it makes sense to stop on a ratio \emph{equal} to 1, in which case we have found an exact maximum. This is what we do in some of our experiments in section \ref{sec:experiments}.}; in sampling mode, we accept $x$ based on a Bernoulli trial of probability $r$.

On line (5), we update the history by recording the trial $x$ and whether it was accepted or not.

If $x$ was rejected (line (6)), then on line (7), we perform a refinement of $q$, based on the principles that we have explained.

\subsection{A connection with A*}
\label{sec:piecewise}

\begin{changedEnv}

\begin{figure}
\centering{\includegraphics[clip=true, draft=false, trim=6cm 5cm 6cm 5cm, scale=.5]{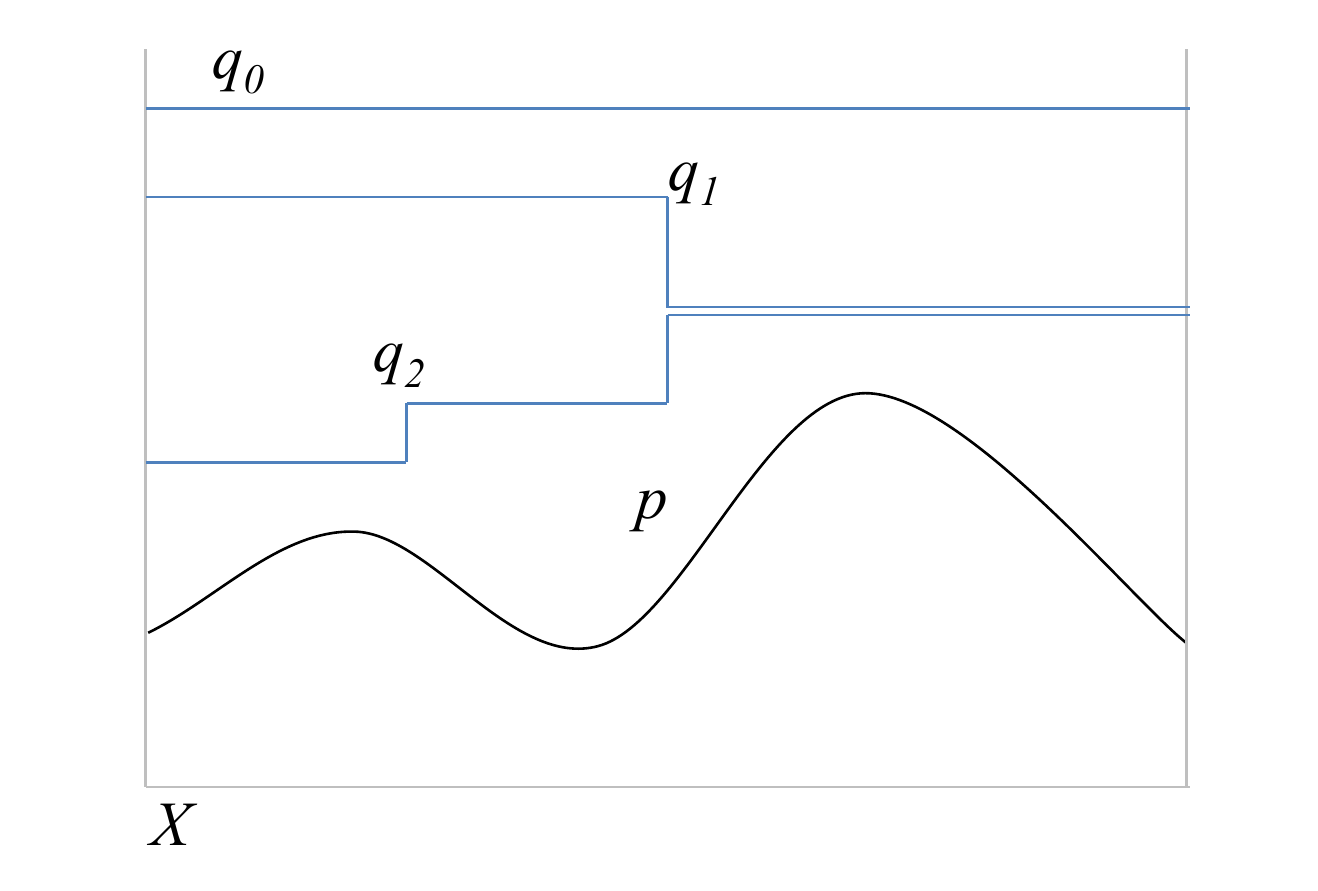}}
\caption{A connection with \Astar.}
	\label{fig:ARSstar1}
\end{figure}

A special case of the \OSstar algorithm, which we call ``\OSstar with piecewise bounds'', shows a deep connection with the classical \Astar optimization algorithm \citep{Hart1968} and is interesting in its own right. Let us first focus on sampling, and let us suppose that $q_0$ represents an initial proposal density, which upper-bounds the target density $p$ over $X$. We start by sampling with $q_0$, and on a first reject somewhere in $X$, we split the set $X$ into two disjoint subsets $X_1,X_2$, obtaining a partition of $X$. 
By using the more precise context provided by this partition, we may be able to improve our upper bound $q_0$ over the whole of $X$ into tighter upper bounds on each of $X_1$ and $X_2$, resulting then in a new upper bound $q_1$ over the whole of $X$. We then sample using $q_1$, and experience at some later time another reject, say on a point in $X_1$; at this point we again partition $X_1$ into two subsets $X_{11}$ and $X_{12}$, tighten the bounds on these subsets, and obtain a refined proposal $q_2$ over $X$; we then iterate the process of building this ``hierarchical partition'' until we reach a certain acceptance-rate threshold.

If we now move our focus to optimization, we see that the refinements that we have just proposed present an analogy to the technique used by \Astar. This is illustrated in Fig.~\ref{fig:ARSstar1}.
In \Astar, we start with a \emph{constant} optimistic bound --- corresponding to our $q_0$ --- for the objective function which is computed at the root of the search tree, which we can assume to be binary. We then expand the two daughters of the root, re-evaluate the optimistic bounds there to new constants, obtaining the piecewise constant proposal $q_1$, and move to the daughter with the highest bound. We continue by expanding at each step the leaf of the partial search tree with the highest optimistic bound (e.g. moving from $q_1$ to $q_2$, etc.).\footnote{\changed{\OSstar, when used in optimization mode, is in fact \emph{strictly more general than \Astar,} for two reasons: (i) it does not assume a piecewise refinement strategy, namely that the refinements follow a hierarchical partition of the space, where a given refinement is limited to a leaf of the current partition,
and (ii) even if such a stategy is followed, it does not assume that the piecewise upper-bounds are constant. Both points will become clearer in the HMM experiments of section \ref{sec:hmms}, where including an higher-order n-gram in $q$ has impact on several regions of $X$ simultaneously, possibly overlapping in complex ways with regions touched by previous refinements; in addition, the impact of a single n-gram is non-constant even in the regions it touches, because it depends of the multiplicity of the n-gram, not only on its presence or absence.}}

We will illustrate \OSstar sampling with (non-constant) piece-wise bounds in the experiments of section \ref{sec:ExperimentsGraphical}.

\end{changedEnv}

\section{Experiments}\label{sec:experiments}

\subsection{HMMs}
\label{sec:hmms}

\def\plm{\p_{\textrm{lm}}}
\def\pobs{\p_{\textrm{obs}}}

\noindent\emph{Note: An extended and more detailed version of these experiments is provided in \citep{Carter2012}}.

\bigskip

The objective in our HMM experiments is to sample a word sequence with density $\pbar(x)$ proportional to $\p(x)=\plm(x)\ \pobs(o|x)$, where $\plm$ is
the probability of the sequence $x$ under an $n$-gram model and $\pobs(o|x)$ is the probability of 
observing the noisy sequence of observations $o$ given that the word sequence is $x$. 
Assuming that the observations depend only on the current state, this probability can be written:
\begin{eqnarray}
p(x) &=& \prod_{i=1}^\ell  \plm(x_i|x^{i-1}_{i-n+1})\ \pobs(o_i|x_i) \enspace.
\label{eq:hmm}
\end{eqnarray}

\paragraph{Approach}

Taking a tri-gram language model for simplicity, let us define $w_3(x_i|x_{i-2} x_{i-1}) =  \plm(x_i|x_{i-2} x_{i-1})\ \pobs(o_i|x_i)$. 
Then consider the
 observation $o$ be fixed, and write $p(x) = \prod_i w_3(x_i|x_{i-2} x_{i-1})$. In optimization/decoding,
 we want to find the argmax of $p(x)$, and in sampling, to sample from $p(x)$. Note that the state space 
associated with $p$ can be huge, as we need to represent explicitly all contexts $(x_{i-2}, x_{i-1})$ in
 the case of a trigram model, and even more contexts for higher-order models.

\sloppy
We define $w_2(x_i| x_{i-1}) = \max_{x_{i-2}} w_3(x_i|x_{i-2} x_{i-1})$, along with 
$w_1(x_i) = \max_{x_{i-1}} w_2(x_i|x_{i-1})$, where the maxima are taken over all possible 
context words in the vocabulary.
These quantities, which can be precomputed efficiently, 
can be seen as optimistic ``max-backoffs'' 
of the trigram  $x_{i-2}^i$, where we have forgotten some part of the context.
Our initial proposal is then $q_{0}(x) = \prod_i w_1(x_i)$. Clearly, for any sequence
$x$ of words, we have $p(x) \leq q_{0}(x)$.  
The state space of $q_0$ is much less costly to represent than that of $p(x)$.
\fussy

\begin{figure}
\includegraphics[clip=true, draft=false, trim=0cm 3.2cm 2.8cm 4cm, scale=.5]{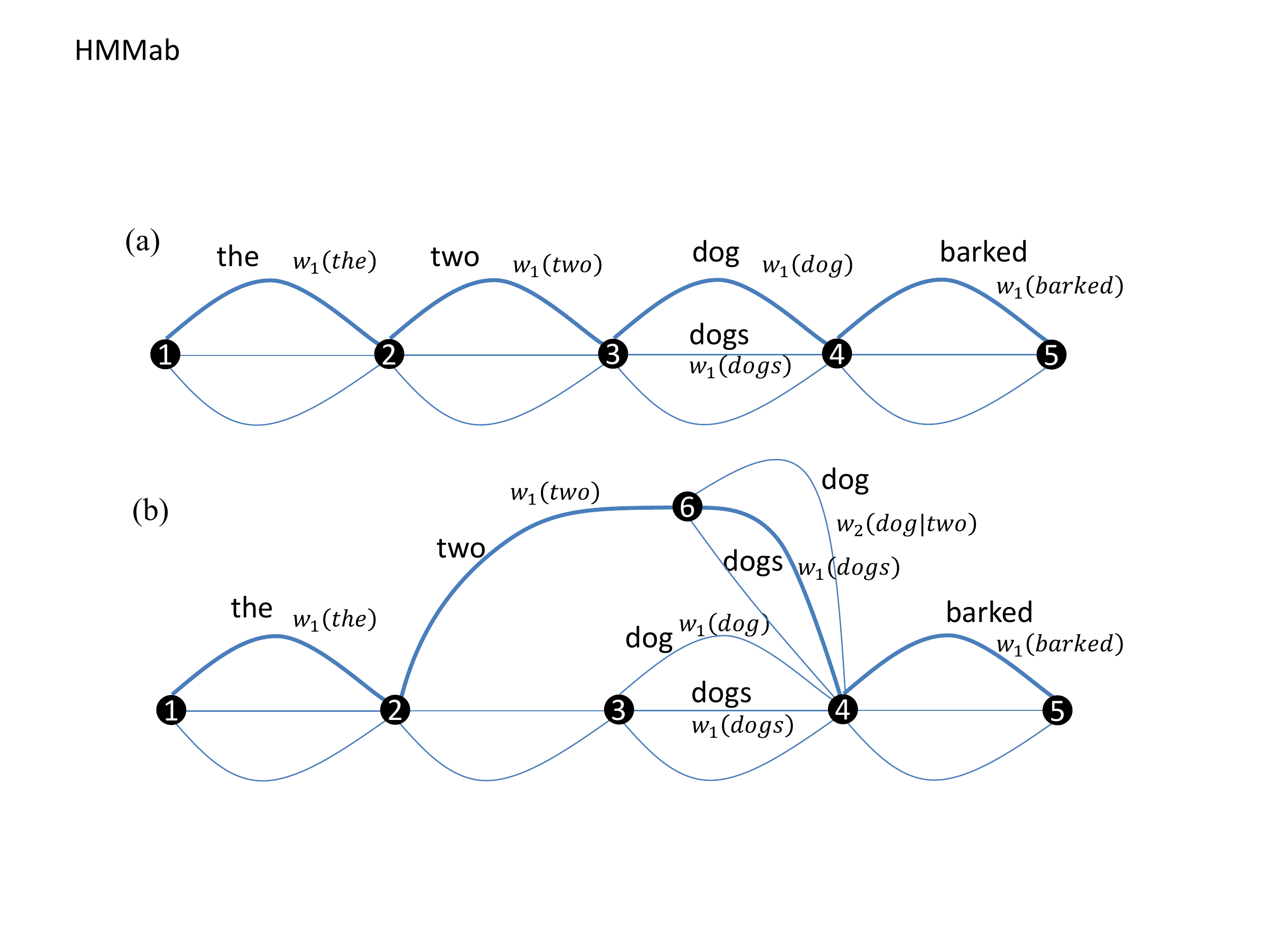}
\caption{{\it An example of an initial q-automaton (a), and its refinement (b).}}
\label{fig:HMMab}
\vspace{-10pt}
\end{figure}

The proposals $q_{t}$, which incorporate n-grams of variable orders, can be represented efficiently through {WFSAs} (weighted FSAs). In Fig.~\ref{fig:HMMab}(a), we show a WFSA representing the initial
proposal $q_0$ corresponding to an example with four observations, which we take to be the acoustic
realizations of the words `the, two, dogs, barked'. The weights on edges correspond 
only to unigram max-backoffs, and thus each state corresponds to a NULL-context. Over
this WFSA, both optimization and sampling can be done efficiently by the standard dynamic
programming techniques (Viterbi~\citep{rabiner}and ``backward filtering-forward sampling''~\citep{Scott2002}), where the forward weights 
associated to states are computed similarly, either in the max-product or in the sum-product semiring.

Consider first sampling, and suppose that the first sample from $q_0$ produces
$x_1 =$ \textit{the two dog barked}, marked with bold edges in the drawing. Now, 
computing the ratio $p(x_1)/q_0(x_1)$ gives a result much smaller than $1$, in part because 
from the viewpoint of the full model $p$, the trigram \textit{the two dog} is 
very unlikely; i.e. the ratio $w_3(dog| the\ two)/w_1(dog)$ is very low. 
Thus, with high probability, $x_1$ is rejected. 
When this is the case, we produce a refined proposal $q_1$, 
represented by the WFSA in Fig.~\ref{fig:HMMab}(b), which takes into account the more
realistic bigram weight $w_2(dog| two)$ by adding a node (node 6) for the context \textit{two}.
We then perform a sampling trial with $q_1$, which this time tends to avoid producing
\textit{dog} in the context of \textit{two}; if we experience a reject later on some sequence $x_2$, we refine again, meaning that we identify an n-gram in $x_2$, which, if we extend its context by one (e.g from a unigram to a bigram or from a bigram to a trigram), accounts for some significant part of the gap between $q_1(x_2)$ and $p(x_2)$. We stop the refinement process when we start observing acceptance rates above a certain fixed threshold.

The case of optimization is similar. Suppose that with $q_0$ the maximum is $x_1 =$ 
\textit{the two dog barked}, then we observe that $p(x_1)$ is lower than $q_0(x_1)$, 
reject $x_1$ and refine $q_0$ into  $q_1$. We stop the process at the point where 
the value of $q_t$, at its maximum $x_{q_{t}}$, is equal to the value of $p$ at $x_{q_{t}}$,
which implies that we have found the maximum for $p$.

\paragraph{Setup}

\def\num{\textrm{num}}
We evaluate our approach on an SMS-message retrieval task. 
Let $N$ be the number of possible words in the vocabulary. 
A latent variable $x\in\{1,\cdots,N\}^\ell$ represents a sentence defined as a sequence of $\ell$ words.
Each word is converted into a sequence of numbers based on a mobile phone numeric keypad, assuming some level of random noise in the conversion. 
The task is then to recover the original message. 

We use the English side of the Europarl corpus~\citep{europarl} for training and test data (1.3 million sentences).
A 5-gram language model is trained using SRILM~\citep{srilm} on 90\% of
the sentences. On the remaining 10\%, we randomly select 100 sequences
for lengths 1 to 10 to obtain 1000 sequences
from which we remove the ones containing numbers, 
obtaining a test set of size 926.

\begin{figure}
\centering{\includegraphics[width=1.2\linewidth,trim=2cm 3cm 0cm 6cm]{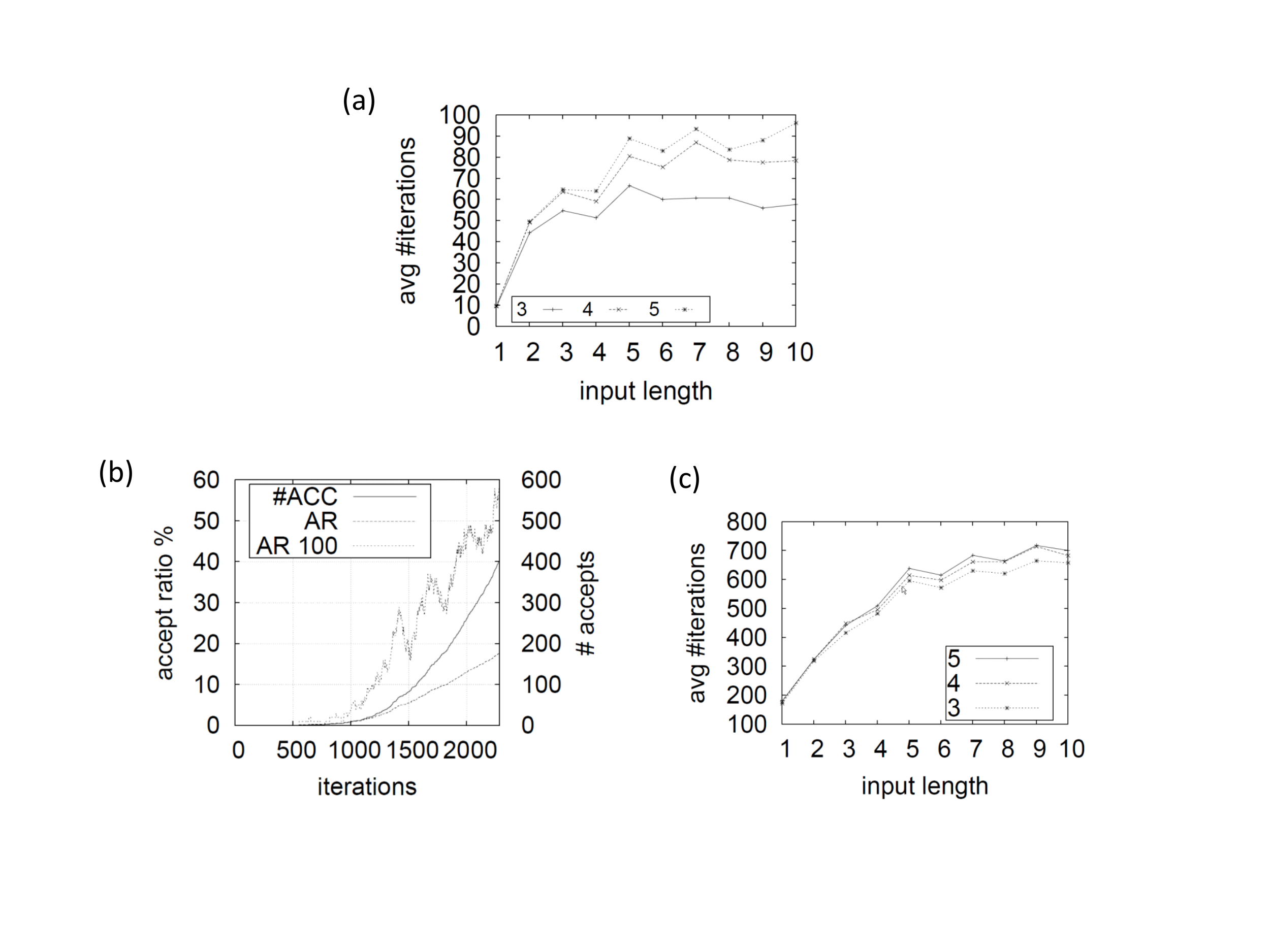}}
\caption{\label{fig:sms-n-iter-states} SMS-retrieval experiment. 
(a): optimization; (b) and (c): sampling.}
\label{fig:sms-decoding}
\end{figure}

\paragraph{Optimization}

We limit the average number of latent tokens in our decoding experiments to 1000.
 In the top plot (a) of Fig.~\ref{fig:sms-decoding} 
we show the number of iterations (running Viterbi then updating $q$) that the
different n-gram models of size 3, 4 and 5 take to do exact decoding of the test-set.
For a fixed sentence length, we can see that decoding with larger n-gram models leads to a sub-linear
increase w.r.t. $n$ in the number of iterations taken.

To demonstrate the reduced nature of our q-automaton, we 
show in Table~\ref{tab:sms-ngram-diff} the distribution of n-grams
in our final model for a specific input sentence of length 10. 
The total number of n-grams in the full model would be ${\sim} 3.0 {\times} 10^{15}$; exact decoding here is not tractable using existing techniques. By comparison, our HMM has only 118 five-grams and 9008 n-grams in total.

\begin{table}[h]\footnotesize
\begin{tabular*}{1\linewidth}{@{\extracolsep{\fill}} c | c | c | c | c | c }
n: & 1 & 2 & 3 & 4 & 5 \\
\hline
q: & 7868 & 615 & 231 & 176 & 118 \\
\end{tabular*}
\caption{\label{tab:sms-ngram-diff} {\it\# of n-grams in our variable-order HMM.}}
\end{table}
\vspace{-10pt}

\paragraph{Sampling}

For the sampling experiments, we limit the number of latent tokens
to 100. We refine our $q$ automaton until we reach 
a certain fixed cumulative acceptance rate (AR). We also compute a rate based only
on the last 100 trials (AR-100), as this tends
to better reflect the current acceptance rate.
In plot (b), at the bottom of Fig.~\ref{fig:sms-decoding}, 
we show a single sampling run using a 5-gram model 
for an example input, and the cumulative \# of accepts (middle curve). 
It took 500 iterations before the first successful
sample from $p$. 

We noted that there is a trade-off between the time needed to compute
the forward probability weights needed for sampling,
and the time it takes to adapt the variable-order HMM. To resolve this, we use batch-updates: making $B$ trials from the same
$q$-automaton, and then updating our model in one step. By doing this,
we noted significant speed-ups in sampling times. Empirically,
we found $B=100$ to be a good value.
In plot (c),
we show the average \# of iterations in our models
once refinements are finished (AR-100=20\%) for different
orders $n$ over different lengths. We note a sub-linear increase in
the number of trials when moving to 
higher $n$; for length${=}$10, and for $n=3,4,5$, average number of trials: 3-1105, \, 4-1238, \, 5-1274.

\subsection{Discrete Probabilistic Graphical Models}
\label{sec:ExperimentsGraphical}

\def\edgeSet{\mathcal{E}}
\def\treeSet{\mathcal{T}}
\def\nodeSet{\mathcal{N}}
\def\neigh{C}

\paragraph{Approach}

The \OSstar approach 
can be applied to exact sampling and optimization on graphical models with loops, where the objective function takes the form of a product of local potentials:
\begin{eqnarray}
p(x) = \prod_{n\in\nodeSet} \psi_n(x) \prod_{e\in\edgeSet} \phi_e(x),
\label{eq:pgm}
\end{eqnarray}
where $(\nodeSet,\edgeSet)$ defines an undirected graph with nodes $\nodeSet$ and \changed{edges} $\edgeSet$. The unary potential functions are denoted by $\psi_n, n\in\nodeSet$ and the binary potentials by $\phi_e, e\in\edgeSet$. Since integrating and sampling from (\ref{eq:pgm}) can be done efficiently for 
trees, we \changed{first determine a spanning tree $\treeSet$ of the graph $\edgeSet$. Let us denote by $\phi_e^{\text{max}}$ and $\phi_e^{\text{min}}$ the maximal and minimal values of the potential for edge $e$. If we define:}
\begin{eqnarray}
q(x) = \prod_{n\in\nodeSet} \psi_n(x) \prod_{e\in\treeSet} \phi_e(x) \prod_{e\in\edgeSet - \treeSet} \phi_e^{\text{max}},
\label{eq:pgm-ub}
\end{eqnarray}
then $q$ is an upper-bound for $p$ over $X$.\footnote{\changed{Any choice of tree produces an upper-bound, but it is advantageous to choose one for which $q$ is as ``close'' as possible to $p$, which we heuristically do by using Prim's algorithm~\citep{Prim1957} for selecting a maximum spanning tree on the graph having weights $\log({\phi_e^{\text{max}}}/{\phi_e^{\text{min}}})$, $e\in\edgeSet$. The intuition is that edges with nearly constant potentials create a small gap between the exact value
$\phi_e(x)$ and the the bound $\phi_e^{\text{max}}$ and can be left outside the tree.}}

To improve this upper bound, we use the \emph{conditioning} idea (see e.g. \citep{Koller2009}, chap. 9.5) which corresponds to partitioning the configuration space $X$. Assume we observe all the possible values $1,2,\cdots,K$ of a node, say $x_i$, having the set of incident edges $\neigh_i$. \changed{The restriction of $p$ to the subspace $X_{i,k} =\{x\in X;x_i=k\}$ can be written as:} 
\begin{eqnarray*}
\changed{p_{i,k}(x)} = \psi_i(k) \underbrace{  \prod_{n\in\nodeSet-\{i\}} \psi_n(x) 
\prod_{e\in\neigh_i} \phi_e(x)
}_{\textrm{unary potentials}}\quad
\underbrace{ \prod_{e\in\edgeSet-\neigh_i} \phi_e(x) }_{\textrm{binary potentials}} ,
\label{eq:pgm-cond}
\end{eqnarray*}
where the binary potentials of edges incident to $x_i$ have now been absorbed into unary potentials associated with neighboring nodes to $x_i$, and where we have used the informal notation $\psi_i(k)$ in an obvious way.
Note that, by doing so, we have eliminated from the graph the edges incident to $x_i$, which may have been participating in loops; in particular, if in equation (\ref{eq:pgm-ub}) we had some edge  $e\in\edgeSet - \treeSet$ incident to $x_i$, then this edge is now absorbed in one of the neighbors of $x_i$, which implies that the maximum $\phi_e^{\text{max}}$ has now been replaced by its exact value, which is necessarily lower, implicitly refining $q$. By conditioning with respect to all the possible values of $x_i$, we obtain $K$ different graphs with the node $x_i$ removed, in other words we have partitioned the event space into $K$ subspaces; \changed{we then define on each such subspace a proposal $q_{i,k}(x)$ as in equation~(\ref{eq:pgm-ub}), which is lower than the restriction of $q$ to $X_{i,k}$.
We then define the refinement $q'$ of $q$ on the whole of $X$ as $q'(x) = \sum_{k} q_{i,k}(x)$, where $q_{i,k}(x)$ is taken to be null for $x\not\in X_{i,k}$. This scheme can be seen to be an instance of $\OSstar$ with piecewise bounds.}

If we repeat this process iteratively based on observed rejections in one of the subspaces, we obtain a hierarchical partition of $X$ which is more fine-grained in some regions of $X$ than in others.
The refinements obtained have the form~(\ref{eq:pgm-ub}), but on reduced graphs in which we have introduced the evidences given by the conditioned variables. While the cardinality of the hierarchical partition may grow exponentially, we can monitor the acceptance rate/computation time ratio, as we will see below.

\paragraph{Ising model experiment} In what follows, we only consider exact sampling, the problem of MAP estimation \changed{(i.e. optimization)} in graphical models having been thoroughly investigated over the past decade~(e.g. \citep{SontagEtAl_uai08}). 
%
%
\changed{One interesting question is to understand the trade-off between improving the acceptance rate and incurring the cost of performing a refinement}. 
\changed{An \emph{a priori} possible policy could be to first refine the proposals up to a certain point and only then sample until we get the required number of samples.
As the experiments will show, there are however two reasons to interleave sampling with refining: the first is that observing rejected samples from $q$ helps to choose the refinements, as argued previously; the second is to have a criterion for stopping the refinement process: the computation of this criterion requires an estimate of the current acceptance rate which can be estimated from the samples.}

We consider a Ising model of 100 binary variables on a 10x10 uniform grid
with unary potentials and binary coupling strengths  drawn according to a centered normal distribution with standard deviation 0.1. 
Note that, \changed{in certain cases}, exact sampling for Ising models can be done in polynomial time~\citep{Ullrich2010}
by using an elegant MCMC approach called \emph{coupling from the past}~\citep{Propp1996}.
It is based on the fact that two properly coupled Markov chains follow exactly the target distribution at the time they coaelesce. However, \changed{applications of this approach rely on certain strong assumptions on the potentials, typically that they are either all attractive or all repulsive~\citep{Craiu2011}}, while we sample models with random positive or negative coupling strengths, making the problem much harder. One interesting extension of our work would be to use these algorithms as proposal distributions for more general models. 

\begin{figure*}
\centering
\includegraphics[width=.9\linewidth,trim=0cm 4cm 0cm 6cm]{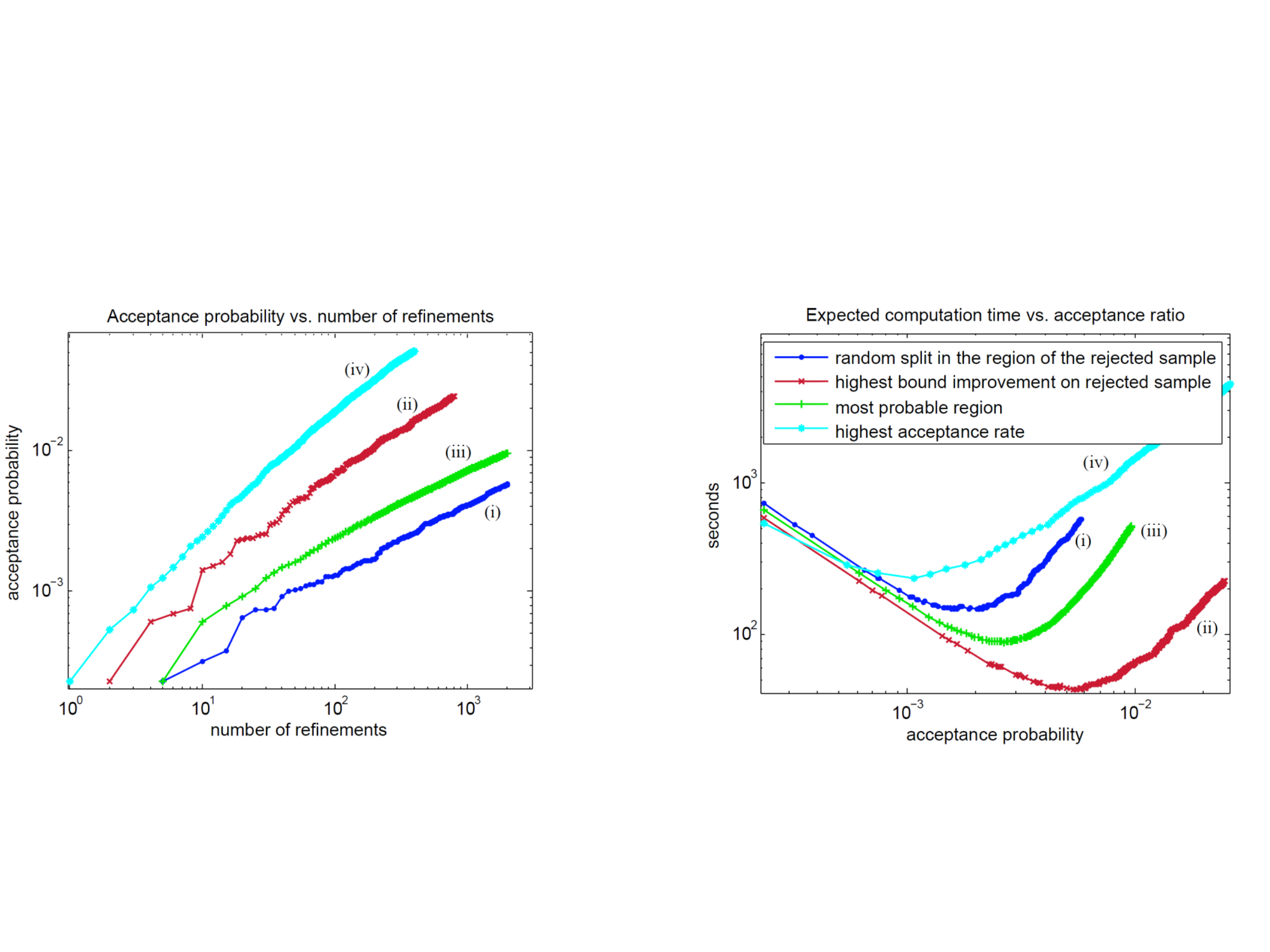}
\caption{Comparison of different refinement policies for the piecewise bound on a 10x10 Ising model grid.}
\label{fig:strategies}
\end{figure*}

The $OS^*$ algorithm was run using 4 different policies
for the choice of the refinements, where a policy is a function that takes as input the current proposal and returns 
a (subspace, conditioning-variable) pair. The first two policies in addition use a rejected sample $x$ as input, while the last two \changed{do not and are deterministic}:
(i)~\emph{random split in the region of the rejected sample} means that once an observation has been rejected, we refine the subspace which contains the sample by conditioning with respect to one of the remaining variables \changed{$x_i$} selected at random;
(ii)~\emph{highest bound improvement on rejected sample} is the same but the \changed{index $i$ of the conditioning variable is selected such that refining on $x_i$ leads to the largest decrease in the value of $q(x)$ (for this specific rejected sample $x$); this is a similar refinement strategy to the one we used in the HMM experiments;} 
(iii)~\emph{most probable region} refines a variable (selected uniformly at random) in the most probable subspace of the piecewise bound $q$; 
(iv)~\emph{highest acceptance rate} is the \changed{most ``ambitious''} greedy policy where the largest reduction of the total mass $q(X)$ of the proposal is identified among all the possible choices of (subspace, conditioning-variable). This has been implemented efficiently by maintaining a priority queue containing all the possible triples (subspace, conditioning-variable, maximal-bound-improvement).
The acceptance rates obtained by following these policies are compared on Figure~\ref{fig:strategies} (left) by running 2000 refinements
with policies (i) and (iii), 800 refinements with policy \changed{(ii)} and 400 refinement with policy (iv).
Results confirm that the best refinement is obtained using the deterministic policy (iv).
The policy (ii) based on rejected samples reaches the same acceptance rates using twice as
many refinements. The other two policies (i) and (iii) are very naive and do not reach a significant 
acceptance rate, even after 400 iterations. Based on these results, one might conclude that the 
policy (iv) is the best. However, we will now see that this is not the case when the computation time is taken into account.

After $T$ trials, the partition function of \changed{$p$, namely $p(X)$,} can be estimated (without bias) by $\hat Z_T\equiv \frac{1}{T}\sum_{t=1}^T r_t\:q_t(X)$
where \changed{the observed accept ratios $\{r_t\}_{t=1}^T$ and refinement weights $\{q(X)_t\}_{t=1}^T$ are used}.
Hence, \changed{$\hat\pi_T = \frac{\hat Z_T}{q_T(X)}$} is an unbiased estimate of the current acceptance rate $\pi_T \equiv p(X)/q_T(X)$.
Note that these estimators are based on \emph{all} the samples $x_t$, accepted or not. Hence, if we decide to stop refining now,
the expected time to obtain $n$ additional exact samples from the target distribution $p$ is
approximately $\frac{n\tau_T^{\textrm{samp}}}{\hat\pi_T}$, where $\tau_T^{\textrm{samp}}$ is the average time to obtain one \changed{trial} from the 
distribution $q_T$. 
If we add the total time $\tau_T^{\textrm{ref}}$ \changed{spent in computing} the set of refinements up to the current refinement $q_T$, we obtain an estimate $\hat\tau^{\textrm{tot}}_T$ of the expected total time to obtain $n$ samples:
\(
\hat\tau^{\textrm{tot}}_T = \frac{n\tau_T^{\textrm{samp}}}{\hat\pi_T} + \tau_T^{\textrm{ref}}\enspace.
\label{eq:total-time}
\)
This quantity computed for $n=1$ sample is plotted against the acceptance rate on Figure~\ref{fig:strategies}~(right). 
For each policy, the expected computation time starts to decrease as the acceptance rate increases; in this regime, 
the refinement time is small compared to the time reduction due to a higher acceptance rate. For large values of the acceptance rate,
the refinement time is no more negligeable, leading to an increase of the total computation time.
We see (Figure~\ref{fig:strategies}~(right)) that the \emph{highest acceptance rate} policy \changed{(iv)}, despite its very good acceptance rate for a fixed number of refinements, requires more time in total than alternative refinements policies. The difference is 
striking: if we look at the minimum of each curve (\changed{which corresponds to the optimal stopping time for the refinements}), it
is at least 10 times faster to use the policy (ii) based on a refinement rule applied only to the rejected sample.
This experiment confirms the benefit of using rejected samples: by adaptively choosing the refinements, we spot the regions of the space that are important to refine much faster than by computing the best possible alternative refinement.

\section{Conclusions and Perspectives}
\label{sec:conclusion}

\changed{In this paper, we have proposed a unified viewpoint for rejection sampling and heuristic optimization, by using functional upper-bounds for both. While in sampling, the upper-bounds are refined by decreasing their integral ($L_1$ norm), in optimization they are refined by decreasing their maximum ($L_\infty$ norm).}
Depending on the problem, several classes of upper bounds can be used. We 
showed that variable-order max-backoff bounds on $n$-gram probabilities gave state-of-the art performances
on the exact decoding of high-order HMMs. For many practical problems, simpler piecewise bounds can be derived, which we illustrated on the example of sampling and decoding on a large tree-width graphical model.
One interesting property of the proposed approach is the 
adaptive nature of the algorithm: the rejected sample is used to quickly choose an effective refinement, 
an approach which can be computational attractive compared to the computation of the $L_p$ norm of all the potential one-step refinements. In the case of graphical model sampling, we showed that this can lead to a speedup factor of an order of magnitude.

The results presented in this paper motivate
further research in the development of domain-specific functional bounds. 
One important extension will be to derive bounds for \emph{agreement-based} models 
$p(.)=p_1(.)\,p_2(.)$ corresponding to the product of two (or more) simple models $p_1$ and $p_2$.
One typical example corresponds to the agreement between an HMM $p_1$ and a probabilistic context-free grammar $p_2$
in order to take into account both the syntactic and the low-level $n$-gram structures of the language \cite{DD-nlp}.

Another research direction is to improve those models that are based on piecewise bounds: 
their quality can be limited since their refinement is always local,
 i.e. they can only improve the bound on a single element of the partition. In the example of graphical model sampling, if we condition on the value of a first variable, say $x_1$, and then further refine the partition $\{x_1=k\}$ by conditioning on the value of a second variable $x_2$, the complementary region of the space $\{x_1\neq k\}$ will not be impacted. Intuitively, the quality of the bound could be improved if we could refine it \emph{jointly} for $x_1$ and $x_2$, in a way analogous to what was done in the context of HMMs, where the incorporation of one higher-order n-gram into the proposal had a global impact on the whole event space.


\clearpage

\bibliographystyle{plainnat}

\end{document}